\documentclass{article} 
\usepackage{iclr2020_conference,times}

\usepackage{amsmath,amsfonts,bm}

\def\eqref#1{equation~\ref{#1}}

\def\1{\bm{1}}

\DeclareMathAlphabet{\mathsfit}{\encodingdefault}{\sfdefault}{m}{sl}
\SetMathAlphabet{\mathsfit}{bold}{\encodingdefault}{\sfdefault}{bx}{n}

\usepackage{hyperref}
\usepackage{url}
\usepackage{latexsym}
\usepackage{graphicx}
\usepackage{framed}
\usepackage{amsfonts}
\usepackage{amsmath}
\usepackage{bbm}
\usepackage{booktabs, tabularx}
\usepackage{subcaption}
\usepackage{comment}

\usepackage{xspace}
\usepackage{tikz}
\usepackage{caption}
\usetikzlibrary{automata,positioning}
\usepackage{verbatim}
\usepackage{multirow}
\usepackage{verbatim}
\usepackage{url}

\newcommand{\blank}{\textsc{[mask]}}
\newcommand{\clstoken}{\textsc{[cls]}}
\newcommand{\bert}{BERT}
\newcommand{\ctx}{\mathbf{x}}
\newcommand{\ack}{RELIC\xspace}
\newcommand{\tac}{TAC-KBP 2010}
\newcommand{\conll}{CoNLL-Aida}

\newcommand{\context}{g}
\newcommand{\concept}{f}

\title{Learning Cross-Context Entity Representations from Text}

\author{\centerline{Jeffrey Ling$^\dagger$\quad Nicholas FitzGerald\quad Zifei Shan\quad Livio Baldini Soares}\\
{\centerline{\bf Thibault F\'evry$^\dagger$\quad David Weiss\quad Tom Kwiatkowski}} \\
Google Research\\
$^\dagger$ Work done as a Google AI Resident.\\ 
\texttt{\{\small jeffreyling,nfitz,zifeis,liviobs,tfevry,djweiss,tomkwiat\}@google.com} \\
}

\iclrfinalcopy 
\begin{document}

\maketitle

\begin{abstract}
Language modeling tasks, in which words, or word-pieces, are predicted on the basis of a local context, have been very effective for learning word embeddings and context dependent representations of phrases.
Motivated by the observation that efforts to code world knowledge into machine readable knowledge bases or human readable encyclopedias tend to be entity-centric, we investigate the use of a fill-in-the-blank task to learn context independent representations of entities from the text contexts in which those entities were mentioned.
We show that large scale training of neural models allows us to learn high quality entity representations, and we demonstrate successful results on four domains: (1) existing entity-level typing benchmarks, including a 64\% error reduction over previous work on TypeNet (Murty et al., 2018); (2) a novel few-shot category reconstruction task; (3) existing entity linking benchmarks, where we match the state-of-the-art on CoNLL-Aida without linking-specific features and obtain a score of 89.8\% on TAC-KBP 2010 without using any alias table, external knowledge base or in domain training data and (4) answering trivia questions, which uniquely identify entities.
Our global entity representations encode fine-grained type categories, such as \emph{Scottish footballers}, and can answer trivia questions such as \emph{Who was the last inmate of Spandau jail in Berlin?}
\end{abstract}

\section{Introduction}

A long term goal of artificial intelligence has been the development and population of an entity-centric representation of human knowledge. 
Efforts have been made to create the knowledge representation with knowledge engineers \citep{lenat1986cyc} or crowdsourcers \citep{bollacker2008freebase}.
However, these methods have relied heavily on human definitions of their ontologies, which are both limited in scope and brittle in nature.
Conversely, due to recent advances in deep learning, we can now learn robust general purpose representations of words \citep{mikolov2013distributed} and contextualized phrases \citep{peters2018deep} directly from large textual corpora.
In particular, we observe that existing methods of building contextualized phrase representations capture a significant amount of local semantic context \citep{devlin2018bert}.
We hypothesize that by learning an \emph{entity encoder} which aggregates all of the textual contexts in which an entity is seen, we should be able to extract and condense general purpose knowledge about that entity.

Consider the following \emph{contexts} in which an entity mention has been replaced a \blank:

\begin{figure}[h]
    \centering
    \footnotesize
    \begin{tabular}{c}
        \dots the second woman in space, 19 years after \blank. \vspace{3pt}\\
        \dots \blank, a Russian factory worker, was the first woman in space \dots \vspace{3pt}\\
         \dots \blank, the first woman in space, entered politics \dots.
    \end{tabular}
    \label{fig:my_label}
\end{figure}
\noindent
As readers, we understand that \emph{first woman in space} is a unique identifier, and we are able to fill in the blank unambiguously. 
The central hypothesis of this paper is that, by matching entities to the contexts in which they are mentioned, we should be able to build a representation for Valentina Tereshkova that encodes the fact that she was the first woman in space, that she was a politician, etc. and that we should be able to use these representations across a wide variety of downstream entity-centric tasks.

We present \ack~(Representations of Entities Learned in Context), a table of independent entity embeddings that have been trained to match fixed length vector representations of the textual context in which those entities have been seen.
We apply \ack~to entity typing (mapping each entity to its properties in an external, curated, ontology); entity linking (identifying which entity is referred to by a textual context), and trivia question answering (retrieving the entity that best answers a question). 
Through these experiments, we show that:

\begin{itemize}
    \item \ack~accurately captures categorical information encoded by human experts in the Freebase and Wikipedia category hierarchies. We demonstrate significant improvements over previous work on established benchmarks, including a 64\% error reduction in the TypeNet low data setting. We also show that given just a few exemplar entities of a given category such as {\it Scottish footballers}~we can use \ack~to recover the remaining entities of that category with good precision.
    \item Using \ack~for entity linking can match state-of-the-art approaches that make use of non-local and non-linguistic information about entities. On the CoNLL-Aida benchmark, \ack~achieves a 94.9\% accuracy, matching the state-of-the-art of \cite{Raiman2018-hm}, despite not using any entity linking-specific features. On the TAC-KBP 2010 benchmark \ack~achieves 89.8\% accuracy, just behind the top ranked system \citep{Raiman2018-hm}, which makes use of external knowledge bases, alias tables, and task-specific hand-engineered features.
    \item \ack~learns better representations of entity properties if it is trained to match just the contexts in which entities are mentioned, and not the surface form of the mention itself. For entity linking, the opposite is true.
    \item We can treat the \ack~embedding matrix as a store of knowledge, and retrieve answers to questions through nearest neighbor search. We show that this approach correctly answers 51\% of the questions in the TriviaQA reading comprehension task \citep{joshi2017triviaqa} despite not using the task's evidence text at inference time. The questions answered correctly by \ack~are surprisingly complex, such as \emph{Who was the last inmate of Spandau jail in Berlin?}

\end{itemize}

\section{Related work}
\label{sec:related}

\paragraph{Entity linking}
The most widely studied entity-level task is entity linking---mapping each entity mention onto a unique entity identifier. 
The Wikification task \citep{ratinov2011local, cheng2013relational}, in particular, is similar to the work presented in this paper, as it requires systems to map mentions to the Wikipedia pages describing the entities mentioned. 
There is significant previous work that makes use of neural context and entity encoders in downstream entity linking systems \citep{sun2015modeling, yamada2016joint, yamada-etal-2017-learning, gupta2017entity, murty2018hierarchical,kolistas2018end}, but that previous work focuses solely on discriminating between entities that match a given mention according to an external alias table. Here we go further in investigating the degree to which \ack~can capture world knowledge about entities.

\paragraph{Mention-level entity typing}
Another well studied task is mention-level entity typing \citep[e.g.][]{ling2012fine, choi2018ultra}. 
In this task, entities are labeled with types that are supported by the immediate textual context. 
For example, given the sentence {\it `Michelle Obama attended her book signing'}, Michelle Obama should be assigned the type {\it author} but not {\it lawyer}. 
Subsequently, mention-level entity typing systems make use of contextualized representations of the entity mention, rather than the global entity representations that we focus on here.

\paragraph{Entity-level typing}
An alternative notion of entity typing is entity-level typing, where each entity should be associated with all of the types supported by a corpus.
\citet{yaghoobzadeh2015corpus} and \citet{murty2018hierarchical} introduce entity-level typing tasks, which we describe more in Section~\ref{sec:entity_typing_results}. 
Entity-level typing is an important task in information extraction, since most common ontologies make use of entity type systems. Such tasks provide a strong method of evaluating learned global representations of entities.

\paragraph{Using knowledge bases} 
There has been a strong line of work in learning representations of entities by building knowledge base embeddings \citep{bordes2011learning, socher2013reasoning, yang2014embedding, toutanova2016compositional, vilnis2018box}, and by jointly embedding knowledge bases and information from textual mentions \citep{riedel2013relation, toutanova2015representing, hu2015entity}.
\citet{das2017question} extended this work to the {\sc spades} fill-in-the-blank task \citep{bisk2016evaluating}, which is a close counterpart to \ack's training setup.
However, we note that all examples in {\sc spades} correspond to a fully connected sub-graph in Freebase~\cite{bollacker2008freebase}. 
Subsequently, the contents 
are very limited in domain and \citet{das2017question} show that it is essential to use the contents of Freebase to do well on this task. 
We consider the unconstrained TriviaQA task \citep{joshi2017triviaqa}, introduced in Section~\ref{sec:trivia-qa}, to be a better evaluation for open domain knowledge representations.

\paragraph{Fill-in-the-blank tasks} 
There has been significant previous work in using fill-in-the-blank losses to learn context independent word representations \citep{mikolov2013distributed}, and context-dependent word and phrase representations \citep{dai2015semi,peters2018deep,radford2018improving,devlin2018bert}.
Cloze-style tasks, in which a system must choose which of a few entities best fill a blanked out span, have also been proposed as a method of evaluating reading comprehension \citep{Hermann:15, long2016leveraging, Onishi:16}. 
For entities, \citet{long2017world} consider a similar fill-in-the-blank task as ours, which they frame as rare entity prediction.
\citet{yamada2016joint} and \citet{yamada-etal-2017-learning} train entity representations using a fill-in-the-blank style loss and a bag-of-words representation of mention contexts. 
\citet{yamada2016joint, yamada-etal-2017-learning}~in particular take an approach that is very similar in motivation to \ack, but which focuses on learning entity representations for use as features in downstream classifiers that model non-linear interactions between a small number of candidate entities. 
In Section~\ref{sec:category_completion}, we show that \cite{yamada-etal-2017-learning}'s entity embeddings are good at capturing broad entity types such as {\it Tennis Player} but less good at capturing more complex compound types such as {\it Scottish Footballers}. 
In Section~\ref{sec:entity_linking_results}, we also show that by performing nearest neighbor search over the 818k entities in the TAC knowledge base, \ack can surpass \citealt{yamada-etal-2017-learning}'s performance on the \tac~entity linking benchmark \citep{Ji10overviewof}. This is despite the fact that \citeauthor{yamada-etal-2017-learning} massively restrict the linking search space with an externally defined alias table, and incorporate task-specific hand-engineered features. On the CoNLL-Aida benchmark, we show that \ack surpasses \citealt{yamada-etal-2017-learning} and matches \citealt{Raiman2018-hm} without using any entity linking-specific features.

\section{Learning from context}\label{sec:learning_setup}

\newcommand{\entstart}{[\textsc{e}_s]}
\newcommand{\entend}{[\textsc{e}_e]}

\subsection{\ack~training input}
Let $\mathcal{E} = \{e_0 \dots e_N\}$ be a predefined set of entities, and let $\mathcal{V} = \{\blank, \entstart, \entend, w_1 \dots w_M\}$ be a vocabulary of words.
A \emph{context} $\ctx = [x_0 \dots  x_l]$ is a sequence of words $x_i \in \mathcal{V}$. 
Each context contains exactly one entity start marker $x_k = \entstart$ and one entity end marker $x_j = \entend$, where $j - k > 1$. 
The sequence of words between these markers, $[x_{k+1} \dots x_{j-1}]$, is the entity mention. 

Our training data is a corpus of (context, entity) pairs $\mathcal{D} = \left[(\ctx_0, y_0)\dots (\ctx_N, y_N)\right]$.
Each $y_i \in \mathcal{E}$ identifies an entity that corresponds to the single entity mention in $\ctx_i$. 
We train \ack~to correctly match the entities in $\mathcal{D}$ to their mentions.
We will experiment with settings where the mentions are unchanged from the original corpus, as well as settings where with some probability $m$ (the \textit{mask rate}) all of the words in the mention have been replaced with the uninformative $\blank$ symbol. We hypothesize that this parameter will play a role in the effectiveness of learned representations in downstream tasks.

For clean training data, we extract our corpus from English Wikipedia\footnote{https://en.wikipedia.org}. See Section~\ref{sec:implementation} for details.

\subsection{context encoder}
We embed each context in $\mathcal{D}$ into a fixed length vector using a Transformer text encoder \citep{vaswani2017attention}, initialized with parameters from the BERT-base model released by \citealt{devlin2018bert}.
All parameters are then trained further using the objective presented below in Section~\ref{sec:ent-ctx-compat}.

We take the output of the Transformer corresponding to the initial $\clstoken$ token in BERT's sequence representation as our context encoding, and we linearly project this into $\mathbb{R}^d$ using a learned weight matrix $W\in \mathbb{R}^{d \times 768}$ to get a context embedding in the same space as our entity embeddings.

\subsection{entity embeddings}
Each entity $e \in \mathcal{E}$ has a unique and abstract Wikidata QID\footnote{https://www.wikidata.org/wiki/Q43649390}.
\ack~maps these unique IDs directly onto a dedicated vector in $\mathbb{R}^d$ via a $\mathcal{|\mathcal{E}|} \times d$ dimensional embedding matrix.
In our experiments, we have a distinct embedding for every concept that has an English Wikipedia page, resulting in 5m entity embeddings overall.

\subsection{\ack~training loss}
\label{sec:ent-ctx-compat}
\ack~optimizes the parameters of the context encoder and entity embedding table to maximize the compatibility between observed (context, entity) pairs. 
Let $\context(\mathbf{x}) \rightarrow \mathbb{R}^d$ be a context encoder, and let $\concept(e) \rightarrow \mathbb{R}^d$ be an embedding function that maps each entity to its $d$ dimensional representation via a lookup operation.
We define a compatibility score between the entity $e$ and the context $\ctx$ as the scaled cosine similarity\footnote{In our experiments, we found cosine similarity to be more effective than dot product.}
\begin{equation}
\label{eqn:entity_score}
      s(\ctx, e) = a \cdot \frac{\context(\ctx)^{\top} \concept(e)}{||\context(\ctx)||||\concept(e)||}
\end{equation}
where the scaling factor $a$ is a learned parameter, following \citet{wang2018additive}.
Now, given a context $\ctx$, the conditional probability that $e$ was the entity seen with $\ctx$ is defined as
\begin{equation}
\label{eqn:entity-prob}
  p(e | \ctx) = \frac{\exp (s(\ctx, e))}{\sum_{e' \in \mathcal{E}} \exp (s(\ctx, e'))}
\end{equation}
and we train \ack~by maximizing the average log probability 
\begin{equation}
    \frac{1}{|\mathcal{D}|} \sum_{(\ctx, y) \in \mathcal{D}} \log p(y| \ctx).
\end{equation}

In practice, the definition of probability in Equation~\ref{eqn:entity-prob} is prohibitively expensive for large $|\mathcal{E}|$ (we use $|\mathcal{E}| \approx 5{\rm M}$). 
Therefore, we use a noise contrastive loss \citep{gutmann2012noise, mnih2013learning}.
We sample $K$ negative entities from a noise distribution $p_{noise}(e)$:
\begin{equation}
e_1', e_2', \ldots, e_K' \sim p_{noise}(e)
\end{equation}
Denoting $e_0' := e$, we then compute our per-example loss using cross entropy:
\begin{equation}
\label{eqn:batch_loss}
    l(s, \ctx, e) = - \log \frac{\exp(s(\ctx, e))}{
    \sum_{j=0}^K \exp(s(\ctx, e_j'))}
\end{equation}


In practice, we train our model with minibatch gradient descent and use all other entries in the batch as negatives.
That is, in a batch of size 4, entities for rows 1, 2, 3 will be used as negatives for row 0. This is roughly equivalent to $p_{noise}(e)$ being proportional to entity frequency.

\section{Experimental setup}
\label{sec:implementation}

To train \ack, we obtain data from the 2018-10-22 dump of English Wikipedia. We take $\mathcal{E}$ to be the set of all entities in Wikipedia (of which there are over 5 million).
For each occurrence of a hyperlink, we take the context as the surrounding sentence, replace all tokens in the anchor text with a single \blank~symbol with probability $m$ (see Section~\ref{sec:masking_ablation} for a discussion of different masking rates) and set the ground truth to be the linked entity.
We limit each context sentence to 128 tokens.
In this way, we collect a high-quality corpus of over 112M (context, entity) pairs.
Note in particular that an entity never co-occurs with text on its own Wikipedia page, since a page will not hyperlink to itself.
We set the entity embedding size to $d = 300$. 

We train the model using TensorFlow \citep{abadi2016tensorflow} with a batch size of 8,192 for 1M steps on Google Cloud TPUs.


\section{Evaluation} \label{sec:evaluation}
We evaluate \ack's ability to: (1) solve the entity linking task without access to any task specific alias tables or features; (2) accurately capture entity properties that have been hand-coded into TypeNet and Wikipedia categories; (3) capture trivia knowledge specific to individual entities.

First we present results on established entity linking and entity typing tasks, to compare \ack's performance to established baselines and we show that the choice of masking strategy (Section~\ref{sec:learning_setup}) has a significant and opposite impact on performance on these tasks. 
We hypothesize that \ack~is approaching an upper bound on established entity-level typing tasks, and we introduce a much harder category completion task that uses \ack~to populate complex Wikipedia categories. We also apply \ack's context encoder and entity embeddings to the task of end-to-end trivia question answering, and we show that this approach can capture more than half of the answers identified by the best existing reading comprehension systems.

\subsection{Entity Linking} \label{sec:entity_linking_results}
\begin{table}[]
    \centering
    \begin{tabular}{|l|c|c|}
    \hline
    System & \conll & \tac \\\hline
    \citealt{Sil2018-qo} & 94.0 & 87.4 \\
    \citealt{yamada2016joint} & 91.5 &  85.5 \\ 
    ~~~~~~~~ - entity linking features & 81.1 & 80.1 \\
    \citealt{yamada-etal-2017-learning}& 94.3 & 87.7 \\
    \citealt{Radhakrishnan2018-zj} & 93.0 & 89.6 \\
    \citealt{Raiman2018-hm} & 94.9 & 90.9 \\
    \ack & 81.9 & 87.5 \\
    \ack + \conll~tuning & 94.9\footnotemark & 89.8 \\
    \hline
    \end{tabular}
    \caption{\ack~achieves comparable precision to best performing dedicated entity-linking systems despite using no external resources or task specific features. When given a standard \conll~alias table and tuned on the \conll~training set, \ack's learned representations match the state-of-the-art DeepType system which relies on the large hand engineered Wikidata knowledge base.}
    \label{tab:entity_linking}
\end{table}
\ack~can be used to directly solve the entity linking problem.\footnotetext{Our finetuned CoNLL result uses the alias table of \cite{pprpershina} at inference time.}
We just need to find the single entity that maximizes the cosine similarity in Equation~\ref{eqn:entity_score} for a given context.
For the entity linking task, we create a context from the document's first 64 tokens as well as the 64 tokens around the mention to be linked.
This choice of context is well suited to the documents in the \conll~and \tac~datasets, since those documents tend to be news articles in which the introduction is particularly information dense.
In Table~\ref{tab:entity_linking}~we show performance for \ack~in two settings. 
First, we report the accuracy for the pure \ack~model with no in-domain tuning.
Then, we report the accuracy for a \ack~model that has been tuned on the CoNLL-Aida training set. 
On the \conll~benchmark, we also adopt a standard alias table \citep{pprpershina}~for this tuned model, as is commonly done in previous entity linking work. 

It is clear that for the \conll~benchmark in-domain tuning is essential.
We hypothesize that this is because of the dataset's bias towards certain types of news content that is very unlike our Wikipedia pre-training data---specifically sports reports. However, when we do adopt the standard CoNLL-Aida training set and alias table, \ack~matches the state of the art on this benchmark, despite using far fewer hand engineered resources (\cite{Raiman2018-hm}~use the large Wikidata knowledge base to create entity representations).
We do not make use of the \tac~training set or alias table, and we observe that \ack~is already competitive without these enhancements\footnote{We do reduce the candidate set from the 5m entities covered by \ack~to the 818k entities in the \tac~knowledge base to avoid ontological misalignment.}

It is significant that \ack~matches the performance of \cite{Raiman2018-hm}, which uses the large hand engineered Wikidata knowledge base to represent entities.
This supports our central hypothesis that it is possible to capture the knowledge that has previously been manually encoded in knowledge bases, using entity embeddings learned from textual contexts alone. In the rest of this section, we will show further support for our hypothesis by recreating parts of the Freebase and Wikipedia ontologies, and by using \ack~to answer trivia questions.

Finally, we believe that \ack's entity linking performance could be boosted even higher through the adoption of commonly used entity linking features.
As shown in Table~\ref{tab:entity_linking}, \cite{yamada2016joint} use a small set of well chosen discrete features to increase the performance of their embedding based approach by 10 points.
These features could be simply integrated into \ack's model, but we consider them to be orthogonal to this paper's investigation of purely learned representations.

\subsection{Entity-level fine typing} \label{sec:entity_typing_results}

We evaluate \ack's ability to capture entity properties on the FIGMENT \citep{yaghoobzadeh2015corpus} and TypeNet \citep{murty2018hierarchical} entity-level fine typing tasks which contain 102 and 1,077 types drawn from the Freebase ontology \citep{bollacker2008freebase}.
The task in both datasets is to predict the set of fine-grained types that apply to a given entity. We train a simple 2-layer feed-forward network that takes as input \ack's embedding $f(e)$ of the entity $e$ and outputs a binary vector indicating which types apply to that entity.

Tables~\ref{tab:figment_results}, \ref{tab:typenet_results} show that \ack significantly outperforms prior results on both datasets.
For FIGMENT, \citet{yaghoobzadeh2018corpus} is an ensemble of several standard representation learning techniques: \texttt{word2vec} skip-gram contexts \citep{mikolov2013distributed}, \emph{structured} skip-gram contexts \citep{ling2015two}, and \texttt{FastText} representations of the entity names \citep{bojanowski2017enriching}.
For TypeNet, \citet{murty2018hierarchical} aggregate mention-level types and train with a structured loss based on the TypeNet hierarchy, but is still outperformed by our flat classifier of binary labels. We expect that including a hierarchical loss is orthogonal to our approach and could improve our results further.

The most striking results in Tables~\ref{tab:figment_results} and \ref{tab:typenet_results} are in the low data settings. 
On the low-data TypeNet setting of \citet{murty2018hierarchical}, \ack~achieves a 63\% error reduction over previous work, while \ack~also matches \citealt{yaghoobzadeh2018corpus}'s results on FIGMENT with 5\% of the training data.
\begin{table}[]
    \centering
    \begin{tabular}{|l|c|c|c|}
    \hline
        System & F1 & P@1 & Acc \\ \hline
        \citealt{yaghoobzadeh2018corpus} &  82.3 & 91.0 & 56.5 \\
        \ack & 87.9 & 94.8 & 68.3 \\
        \ack with 5\% of FIGMENT training data & 83.3 & 90.9 & 59.3 \\ \hline
    \end{tabular}
    \caption{Performance on FIGMENT. We report P@1 (proportion of entities whose top ranked types are correct), Micro F1 aggregated over all (entity, type) compatibility decisions, and overall accuracy of entity labeling decisions. \ack outperforms prior work, even with only 5\% of the training data.}
    \label{tab:figment_results}
\end{table}

\begin{table}[]
    \centering
    \begin{tabular}{|l|c|c|}
    \hline
    System & TypeNet & TypeNet - Low Data (5\%) \\\hline
    \citealt{murty2018hierarchical} & 78.6 & 58.8 \\
    \ack & 90.1 & 85.3 \\\hline
    \end{tabular}
    \caption{Mean Average Precision on TypeNet tasks. \ack's gains are particularly striking in the low data setting from \cite{murty2018hierarchical}.}
    \label{tab:typenet_results}
\end{table}

\subsection{Effect of masking} \label{sec:masking_ablation}

\begin{figure}[t!]
    \centering
    \begin{minipage}{.45\textwidth}
        \centering
        \includegraphics[scale=0.3]{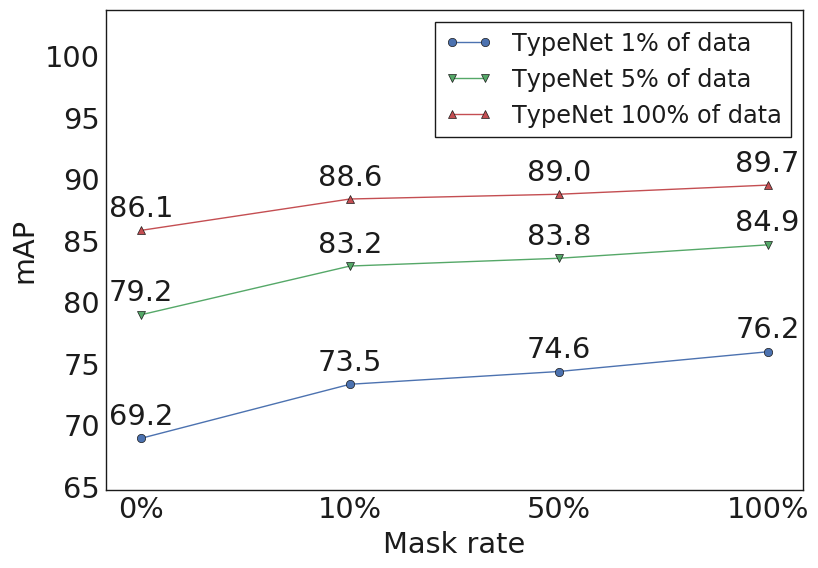}
        \caption{TypeNet entity-level typing mAP on the development set for \ack models trained with different masking rates. A higher mask rate leads to better performance, both in low and high-data situations.}
        \label{fig:typenet_mask_rates}
    \end{minipage}~~~~~~~~~~~~~~~~~~~~~
    \begin{minipage}{0.45\textwidth}
        \centering
        \includegraphics[scale=0.3]{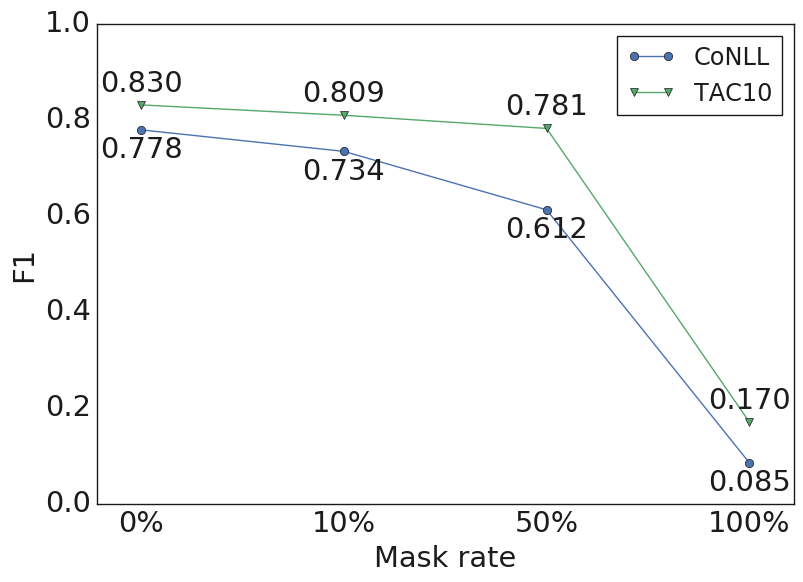}
        \caption{Entity linking accuracy for \ack models trained with different masking rates. No alias table nor in-domain fine-tuning is used. Higher mask rates lead to worse downstream performance in entity-linking tasks.}
        \label{fig:el_mask_rates}
    \end{minipage}
\end{figure}

In Section~\ref{sec:learning_setup} we introduced the concept of masking entity mentions, and predicting on the basis of the context in which they are discussed, not the manner in which they are named. 
Figures~\ref{fig:typenet_mask_rates} and \ref{fig:el_mask_rates} show the effect of training \ack~with different mask rates. 
It is clear that masking mentions during training is beneficial for entity typing tasks, but detrimental for entity linking. 
This is in accordance with our intuitions. Modeling mention surface forms is essential for linking, since these mentions are given at test time and names are extremely discriminative. However, once the mention is known the model only needs to distinguish between different entities with the same name (e.g. {\it President Washington, University of Washington, Washington State}) and this distinction rarely requires deep knowledge of each entity's properties. Subsequently, our best typing models are those that are forced to capture more of the context in which each entity is mentioned, because they are not allowed to rely on the mention itself.
The divergence between the trends in Figures~\ref{fig:typenet_mask_rates} and \ref{fig:el_mask_rates}~suggests that there may not be one set of entity embeddings that are optimum for all tasks. However, we would like to point out that that a mask rate of 10\%, \ack~nears optimum performance on most tasks. The optimum mask rate is an open research question, that will likely depend on entity frequency as well as other data statistics.

\subsection{Few-shot category completion} \label{sec:category_completion}
The entity-level typing tasks discussed above involve an in-domain training step. 
Furthermore, due to the incompleteness of the the FIGMENT and TypeNet type systems, we also believe that \ack's performance is approaching the upper bound on both of these supervised tasks.
Therefore, to properly measure \ack's ability to capture complex types from fill-in-the-blank training alone, we propose:

\begin{enumerate}
\item a new category completion task that does not involve any task specific optimization,
\item a new Wikipedia category based evaluation set that contains much more complex compound types, such as {\it Scottish footballers},
\end{enumerate}

We use this new task to compare \ack~to the embeddings learned by \citealt{yamada-etal-2017-learning}.

In the new category completion task, we represent each category by randomly sampling three exemplar entities, and calculating the centroid of their \ack~embeddings.
We then rank all other entities according to their dot-product with this centroid, and report the mean average precision (MAP) of the resultant ranking.

\begin{table}[]
    \centering
    \begin{tabular}{|l||c|c||c|c|}
    \hline
    & \multicolumn{2}{c||}{Yamada Subset} & \multicolumn{2}{c|}{All Entities} \\ \hline
    & TypeNet & Wikipedia & TypeNet & Wikipedia \\ \hline
    \# Entities & 291,663 & 707,588 & 323,347 & 3,667,933 \\ \hline
    Random & 2.7 & 0.1 & 2.5 & 0.1 \\
    \citealt{yamada-etal-2017-learning} & 25.9 & 8.0 & -- & -- \\
    \ack & 27.8 & 21.0 & 29.3 & 13.8 \\ \hline
    \end{tabular}
    \caption{Mean average precision on exemplar-based category completion (Section~\ref{sec:category_completion}).
    The Yamada subset is filtered to only contain entities that are covered by \citealt{yamada-etal-2017-learning}, and categories are filtered to those
    which contain at least 300 entities (131 categories).
    For the "All Entities" setting, we use all Wikipedia entities covered by \ack, and filter to categories which contain at least 1000 entities (1083 categories).
    The embeddings learned by \citealt{yamada-etal-2017-learning} are competitive with \ack~on the task of populating TypeNet categories, but they are much worse at capturing the complex, and compound, typing information present in Wikipedia categories.}
    \label{tab:category_completion}
\end{table}

First, we apply this evaluation to the TypeNet type system introduced in \citep{murty2018hierarchical}.
These types are well-curated, but tend to represent high-level categories.
To measure the degree to which our entity embeddings capture finer grained type information, we construct an aditional dataset based on Wikipedia categories\footnote{We use the Yago 3.1~\citep{mahdisoltani2013yago3} dump of extracted categories that cover at least 1,000 entities, resulting in 1,083 categories.}.
These tend to be compound types, such as {\it Actresses from London}, which capture many aspects of an entity---in this case gender, profession, and place of birth.

From Table~\ref{tab:category_completion} we can see that the embeddings introduced by \citealt{yamada-etal-2017-learning} approach \ack's  performance on the TypeNet completion task, but they significantly underperform \ack~in completing the more complex Wikipedia categories.
Figure~\ref{tab:category_preds} shows example reconstructions for randomly sampled Wikipedia categories, two from TypeNet and three from Wikipedia. 
Both models achieve high precision on TypeNet categories, but on the finer-grained Wikipeida categories, the \citet{yamada-etal-2017-learning} model tends to
produce more broadly-related entites, whereas the \ack embeddings capture entities which are much closer to the exemplars.
In fact, we identify several false negatives in these examples.



\subsection{Trivia question answering}
\label{sec:trivia-qa}


\begin{table}[]
    \small
    \centering
    \begin{tabular}{|l|c|c|}
    \hline
          & Open-domain Unfiltered & Verified Web \\\hline
    Classifier baseline \citep{joshi2017triviaqa} & ---  & 30.2 \\
    SLQA \citep{wang2018multi}                    & ---  & 82.4 \\
    \hline
    \ack                                          & 35.7 & 51.2 \\ 
    ORQA \citep{lee-etal-2019-latent}             & 45.0 & ---  \\
    \hline
    \end{tabular}
    \caption{Answer exact match on TriviaQA. \ack's~fast nearest neighbor search over entities achieves 80\% of the performance of ORQA, which runs a BERT-based reading comprehesion model over multiple retrieved evidence passages. Unlike ORQA and \ack, the classifier baseline and SLQA have access to a single evidence document that is known to contain the answer. As a result they are solving a much easier task. } \label{tab:triviaqa_results}
\end{table}

Our final experiment tests \ack's ability to answer trivia questions -- which can be considered high precision categories that only apply to a single entity -- using retrieval of encoded entities.
TriviaQA \citep{joshi2017triviaqa} is a question-answering dataset containing questions sourced from trivia websites, and the answers are usually entities with Wikipedia pages. The standard TriviaQA setup is a reading comprehension task, where answers are extracted from evidence documents.
Here, we answer questions in TriviaQA \emph{without access to any evidence documents at test time.}

\paragraph{Model and training} Given a question, we apply the context encoder $\context$ from Section~\ref{sec:ent-ctx-compat}, and retrieve 1 out of 5M entities using cosine similarity.
For training, we initialize both $\context$ and $\concept$ from \ack training. We tune only $\context$'s parameters by optimizing the loss in Equation~\ref{eqn:batch_loss} applied to (question, answer entity) pairs, rather than the (context, entity) pairs seen during \ack's training.

\paragraph{Results}



TriviaQA results are shown in Table~\ref{tab:triviaqa_results}, and randomly sampled \ack~predictions are illustrated in Figure~\ref{tab:triviaqa_preds}.
All systems other than \ack~in Table~\ref{tab:triviaqa_results}~have access to evidence text at inference time.
In the open domain unfiltered setting, ORQA \citep{lee-etal-2019-latent} retrieves this text from a cache of Wikipedia. In the more standard verified-web reading comprehension task, the classifier baseline and SLQA are provided with a single document that is known to contain the answer. 

We consider ORQA to be the most relevant point of comparison for \ack. We observe that the retrieve-then-read approach taken by ORQA outperforms the direct answer retrieval approach taken by \ack.
However, ORQA runs a BERT based reading comprehension model over multiple evidence passages at inference time and we are encouraged to see that \ack's much faster nearest neighbor lookup captures 80\% of ORQA's performance.

It is also significant that \ack~outperforms \cite{joshi2017triviaqa}'s reading comprehension baseline by 20 points, despite the fact that the baseline has access to a single document that is known to contain the answer. However, \ack~is still far behind the reading comprehension upper bound set by \cite{wang2018multi} and there is a long way to go before \ack's embeddings can capture all of the facts that can be identified by question-dependent inference time reading.

\begin{figure*}[]
    \begin{subtable}[b]{\textwidth}
    \scriptsize
    \centering
    \begin{tabular}{|p{0.2\textwidth}|p{0.35\textwidth}|p{0.35\textwidth}|}
        \toprule
        \hline
        \multicolumn{1}{|c|}{Category and Exemplars} & \multicolumn{1}{c|}{\citealt{yamada-etal-2017-learning}} & \multicolumn{1}{c|}{\ack} \\ \hline
        \texttt{tennis\_player} \newline David Goffin \newline Yves Allegro \newline Flavia Pennatta & 1. \textbf{Ekaterina Makarova} \newline 2. \textbf{Vera Zvonareva} \newline 3. \textbf{Flavia Pennetta} \newline 4. \textbf{Max Mirnyi} \newline 5. \textbf{Lisa Raymond} \newline AP=71.87 & 1. \textbf{Prakash Amritraj} \newline 2. \textbf{Marco Chiudinelli} \newline 3. \textbf{Marc Gicquel} \newline 4. \textbf{Marius Copil} \newline 5. \textbf{Benjamin Balleret} \newline AP=56.87 \\ 
        \hline
        \texttt{exhibition\_producer} \newline Toledo Museum of Art \newline Egyptian Museum \newline San Jose Museum of Art &  1. \textbf{Smithsonian American Art Museum} \newline 2. \textbf{Honolulu Museum of Art} \newline 3. \textbf{Brooklyn Museum} \newline 4. \textbf{Whitney Museum of American Art} \newline 5. \textbf{Hirshhorn Museum and Sculpture Garden} \newline AP=38.41 & 1. \textbf{Cleveland Museum of Art} \newline 2. \textbf{Smithsonian American Art Museum} \newline 3. \textbf{Indianapolis Museum of Art} \newline 4. \textbf{Cincinnati Art Museum} \newline 5. \textbf{Museum of Fine Arts, Boston} \newline AP=52.52 \\ 
        \hline
        \hline
        \texttt{Scottish footballers} \newline Pat Crerand \newline Gerry Britton \newline Jim McLean &  1. Ayr United F.C. \newline 2. Clyde F.C. \newline 3. Scottish League Cup \newline 4. Stranraer F.C. \newline 5. Arbroath F.C. \newline AP=4.57 & 1. \textbf{Tommy Callaghan} \newline 2. \textbf{Gordon Wallace} \newline 3. \emph{David White}** \newline 4. Davie Dodds \newline 5. \textbf{John Coughlin} \newline AP=67.10 \\ 
        \hline
        \texttt{Number One Singles in Germany} \newline Lady Marmalade \newline Just Give Me a Reason \newline I'd Do Anything for Love (But I Won't Do That) &  1. Billboard Hot 100 \newline 2.  Grammy Award for Best Female Pop Vocal Performance \newline 3. Dance Club Songs \newline 4. Pop 100 \newline 5. Hot Latin Songs \newline AP=4.14 & 1. Try (Pink song) \newline 2. \emph{Whataya Want from Me}** \newline 3. Fuckin' Perfect \newline 4. Beautiful (Christina Aguilra song) \newline 5. Raise Your Glass \newline AP=4.59 \\ 
        \hline
        \texttt{2010 Albums} \newline This Is the Warning \newline Tin Can Trust \newline Bionic (Christina Aguilera album) &  1. FAA airport categories \newline 2. Rugby league county cups \newline 3. Digital Songs \newline 4. Country Airplay \newline 5. Swiss federal election, 2007 \newline AP=0.04 & 1. \emph{All I Want Is You}** \newline 2. Don't Mess with the Dragon \newline 3. \emph{Believe (Orianthi album)}** \newline 4. Sci-Fi Crimes \newline 5. \textbf{Interpol} \newline AP=6.78 \\ 
        \hline
        \bottomrule

    \end{tabular}
    \caption{Top 5 predictions for a set of randomly selected categories, given 3 exemplars. The first two categories come from TypeNet, and the second two from our Wikipedia categorization dataset. Correct predictions are bolded. Predictions which are judged by the authors to be false negatives (predictions which properly belong to the target category) are indicated with asterisks**.}
    \label{tab:category_preds}
    \end{subtable}
    
    \vspace{0.2cm}

\begin{subtable}[t]{\textwidth}
    \begin{framed}
    \small
        \textbf{\underline{Q:}} Who was the last inmate of Spandau jail in Berlin? \\
        \textbf{\underline{A:}} \textbf{1. Rudolf Hess} 2. Adolf Hitler 3. Hermann Göring 4. Heinrich Himmler 5. Ernst Röhm \\
       
        \textbf{\underline{Q:}} Which fashionable London thoroughfare, about three quarters of a mile (1.2 km) long, runs from Hyde Park Corner to Marble Arch, along the length of the eastern side of Hyde Park? \\
        \textbf{\underline{A:}} \textbf{1. Park Lane} 2. Piccadilly 3. Knightsbridge 4. Leicester Square 5. Tottenham Court Road \\
    
        \textbf{\underline{Q:}} In which Lake District town would you find the Cumberland Pencil Museum? \\
        \textbf{\underline{A:}} \textbf{1. Keswick} 2. Hawkshead 3. Grasmere 4. Cockermouth 5. Ambleside \\
  
        \textbf{\underline{Q:}} The Wimbledon tennis tournament is held at which tennis club in London? \\
        \textbf{\underline{A:}} 1. Queen's Club \textbf{2. All England Lawn Tennis and Croquet Club} 3. Wimbledon Championships 4. Stade Roland-Garros 5. Wentworth Club 
    \end{framed}
    \caption{TriviaQA predictions from retrieval. Questions are randomly sampled, and top 5 ranking answers are shown. Correct answer in bold. Note that even when the model is wrong, the predictions are all of the correct type.}
    \label{tab:triviaqa_preds}
    \end{subtable}
    
    \caption{Random example predictions drawn from category completion, and TriviaQA tasks.}

\end{figure*}

\section{Conclusion}
\label{sec:conclusion}

In this paper, we demonstrated that the \ack fill-in-the-blank task allows us to learn context independent representations of entities with their own latent ontology.
We show successful entity-level typing results on FIGMENT \citep{yaghoobzadeh2015corpus} and TypeNet \citep{murty2018hierarchical}, even when only training on a small fraction of the task-specific training data.
We then introduce a novel few-shot category reconstruction task and when comparing to \citet{yamada-etal-2017-learning}, we found that \ack~is better able to capture complex compound types.  
Our method also proves successful for entity linking, where we match the state of the art on CoNLL-Aida despite not using linking-specific features and fare similarly to the best system on \tac~despite not using an alias table, any external knowledge bases, linking-specific features or even in-domain training data.
Finally, we show that our \ack~embeddings can be used to answer trivia questions directly, without access to any evidence documents. 
We encourage researchers to further explore the properties of our entity representations and BERT context encoder, which we will release publicly.



\clearpage

\bibliography{iclr2020_conference}

\begin{thebibliography}{49}
\providecommand{\natexlab}[1]{#1}
\providecommand{\url}[1]{\texttt{#1}}
\expandafter\ifx\csname urlstyle\endcsname\relax
  \providecommand{\doi}[1]{doi: #1}\else
  \providecommand{\doi}{doi: \begingroup \urlstyle{rm}\Url}\fi

\bibitem[Abadi et~al.(2016)Abadi, Barham, Chen, Chen, Davis, Dean, Devin,
  Ghemawat, Irving, Isard, et~al.]{abadi2016tensorflow}
Mart{\'\i}n Abadi, Paul Barham, Jianmin Chen, Zhifeng Chen, Andy Davis, Jeffrey
  Dean, Matthieu Devin, Sanjay Ghemawat, Geoffrey Irving, Michael Isard, et~al.
\newblock Tensorflow: A system for large-scale machine learning.
\newblock In \emph{OSDI}, 2016.

\bibitem[Bisk et~al.(2016)Bisk, Reddy, Blitzer, Hockenmaier, and
  Steedman]{bisk2016evaluating}
Yonatan Bisk, Siva Reddy, John Blitzer, Julia Hockenmaier, and Mark Steedman.
\newblock Evaluating induced ccg parsers on grounded semantic parsing.
\newblock In \emph{EMNLP}, 2016.

\bibitem[Bojanowski et~al.(2017)Bojanowski, Grave, Joulin, and
  Mikolov]{bojanowski2017enriching}
Piotr Bojanowski, Edouard Grave, Armand Joulin, and Tomas Mikolov.
\newblock Enriching word vectors with subword information.
\newblock \emph{TACL}, 2017.

\bibitem[Bollacker et~al.(2008)Bollacker, Evans, Paritosh, Sturge, and
  Taylor]{bollacker2008freebase}
Kurt Bollacker, Colin Evans, Praveen Paritosh, Tim Sturge, and Jamie Taylor.
\newblock Freebase: a collaboratively created graph database for structuring
  human knowledge.
\newblock In \emph{ACM SIGMOD international conference on Management of data},
  2008.

\bibitem[Bordes et~al.(2011)Bordes, Weston, Collobert, and
  Bengio]{bordes2011learning}
Antoine Bordes, Jason Weston, Ronan Collobert, and Yoshua Bengio.
\newblock Learning structured embeddings of knowledge bases.
\newblock In \emph{AAAI}, 2011.

\bibitem[Cheng \& Roth(2013)Cheng and Roth]{cheng2013relational}
Xiao Cheng and Dan Roth.
\newblock Relational inference for wikification.
\newblock In \emph{EMNLP}, 2013.

\bibitem[Choi et~al.(2018)Choi, Levy, Choi, and Zettlemoyer]{choi2018ultra}
Eunsol Choi, Omer Levy, Yejin Choi, and Luke Zettlemoyer.
\newblock Ultra-fine entity typing.
\newblock In \emph{ACL}, 2018.

\bibitem[Dai \& Le(2015)Dai and Le]{dai2015semi}
Andrew~M Dai and Quoc~V Le.
\newblock Semi-supervised sequence learning.
\newblock In \emph{NIPS}, 2015.

\bibitem[Das et~al.(2017)Das, Zaheer, Reddy, and McCallum]{das2017question}
Rajarshi Das, Manzil Zaheer, Siva Reddy, and Andrew McCallum.
\newblock Question answering on knowledge bases and text using universal schema
  and memory networks.
\newblock In \emph{ACL}, 2017.

\bibitem[Devlin et~al.(2019)Devlin, Chang, Lee, and Toutanova]{devlin2018bert}
Jacob Devlin, Ming-Wei Chang, Kenton Lee, and Kristina Toutanova.
\newblock Bert: Pre-training of deep bidirectional transformers for language
  understanding.
\newblock In \emph{NAACL}, 2019.

\bibitem[Gupta et~al.(2017)Gupta, Singh, and Roth]{gupta2017entity}
Nitish Gupta, Sameer Singh, and Dan Roth.
\newblock Entity linking via joint encoding of types, descriptions, and
  context.
\newblock In \emph{EMNLP}, 2017.

\bibitem[Gutmann \& Hyv{\"a}rinen(2012)Gutmann and
  Hyv{\"a}rinen]{gutmann2012noise}
Michael~U Gutmann and Aapo Hyv{\"a}rinen.
\newblock Noise-contrastive estimation of unnormalized statistical models, with
  applications to natural image statistics.
\newblock \emph{JMLR}, 13\penalty0 (Feb):\penalty0 307--361, 2012.

\bibitem[Hermann et~al.(2015)Hermann, Ko\v{c}isk\'{y}, Grefenstette, Espeholt,
  Kay, Suleyman, and Blunsom]{Hermann:15}
Karl~Moritz Hermann, Tom\'{a}\v{s} Ko\v{c}isk\'{y}, Edward Grefenstette, Lasse
  Espeholt, Will Kay, Mustafa Suleyman, and Phil Blunsom.
\newblock Teaching machines to read and comprehend.
\newblock In \emph{NIPS}, 2015.

\bibitem[Hu et~al.(2015)Hu, Huang, Deng, Gao, and Xing]{hu2015entity}
Zhiting Hu, Poyao Huang, Yuntian Deng, Yingkai Gao, and Eric Xing.
\newblock Entity hierarchy embedding.
\newblock In \emph{ACL}, 2015.

\bibitem[Ji et~al.(2010)Ji, Grishman, Dang, Griffitt, and
  Ellis]{Ji10overviewof}
Heng Ji, Ralph Grishman, Hoa~Trang Dang, Kira Griffitt, and Joe Ellis.
\newblock Overview of the tac 2010 knowledge base population track.
\newblock In \emph{TAC}, 2010.

\bibitem[Joshi et~al.(2017)Joshi, Choi, Weld, and
  Zettlemoyer]{joshi2017triviaqa}
Mandar Joshi, Eunsol Choi, Daniel Weld, and Luke Zettlemoyer.
\newblock Triviaqa: A large scale distantly supervised challenge dataset for
  reading comprehension.
\newblock In \emph{ACL}, 2017.

\bibitem[Kolitsas et~al.(2018)Kolitsas, Ganea, and Hofmann]{kolistas2018end}
Nikolaos Kolitsas, Octavian{-}Eugen Ganea, and Thomas Hofmann.
\newblock End-to-end neural entity linking.
\newblock \emph{CoRR}, 2018.

\bibitem[Lee et~al.(2019)Lee, Chang, and Toutanova]{lee-etal-2019-latent}
Kenton Lee, Ming-Wei Chang, and Kristina Toutanova.
\newblock Latent retrieval for weakly supervised open domain question
  answering.
\newblock In \emph{ACL}, 2019.

\bibitem[Lenat et~al.(1986)Lenat, Prakash, and Shepherd]{lenat1986cyc}
Douglas Lenat, Mayank Prakash, and Mary Shepherd.
\newblock Cyc: Using common sense knowledge to overcome brittleness and
  knowledge acquisition bottlenecks.
\newblock \emph{AI Magazine}, 6:\penalty0 65--85, 12 1986.

\bibitem[Ling et~al.(2015)Ling, Dyer, Black, and Trancoso]{ling2015two}
Wang Ling, Chris Dyer, Alan~W Black, and Isabel Trancoso.
\newblock Two/too simple adaptations of word2vec for syntax problems.
\newblock In \emph{NAACL}, 2015.

\bibitem[Ling \& Weld(2012)Ling and Weld]{ling2012fine}
Xiao Ling and Daniel~S Weld.
\newblock Fine-grained entity recognition.
\newblock In \emph{AAAI}, 2012.

\bibitem[Long et~al.(2016)Long, Lowe, Cheung, and Precup]{long2016leveraging}
Teng Long, Ryan Lowe, Jackie Chi~Kit Cheung, and Doina Precup.
\newblock Leveraging lexical resources for learning entity embeddings in
  multi-relational data.
\newblock In \emph{ACL}, 2016.

\bibitem[Long et~al.(2017)Long, Bengio, Lowe, Cheung, and
  Precup]{long2017world}
Teng Long, Emmanuel Bengio, Ryan Lowe, Jackie Chi~Kit Cheung, and Doina Precup.
\newblock World knowledge for reading comprehension: Rare entity prediction
  with hierarchical lstms using external descriptions.
\newblock In \emph{EMNLP}, 2017.

\bibitem[Mahdisoltani et~al.(2013)Mahdisoltani, Biega, and
  Suchanek]{mahdisoltani2013yago3}
Farzaneh Mahdisoltani, Joanna Biega, and Fabian~M Suchanek.
\newblock Yago3: A knowledge base from multilingual wikipedias.
\newblock In \emph{CIDR}, 2013.

\bibitem[Mikolov et~al.(2013)Mikolov, Sutskever, Chen, Corrado, and
  Dean]{mikolov2013distributed}
Tomas Mikolov, Ilya Sutskever, Kai Chen, Greg~S Corrado, and Jeff Dean.
\newblock Distributed representations of words and phrases and their
  compositionality.
\newblock In \emph{NIPS}, 2013.

\bibitem[Mnih \& Kavukcuoglu(2013)Mnih and Kavukcuoglu]{mnih2013learning}
Andriy Mnih and Koray Kavukcuoglu.
\newblock Learning word embeddings efficiently with noise-contrastive
  estimation.
\newblock In \emph{NIPS}, 2013.

\bibitem[Murty et~al.(2018)Murty, Verga, Vilnis, Radovanovic, and
  McCallum]{murty2018hierarchical}
Shikhar Murty, Patrick Verga, Luke Vilnis, Irena Radovanovic, and Andrew
  McCallum.
\newblock Hierarchical losses and new resources for fine-grained entity typing
  and linking.
\newblock In \emph{ACL}, 2018.

\bibitem[Onishi et~al.(2016)Onishi, Wang, Bansal, Gimpel, and
  McAllester]{Onishi:16}
Takeshi Onishi, Hai Wang, Mohit Bansal, Kevin Gimpel, and David McAllester.
\newblock Who did what: A large-scale person-centered cloze dataset.
\newblock In \emph{EMNLP}, 2016.

\bibitem[Pershina et~al.(2015)Pershina, He, and Grishman]{pprpershina}
Maria Pershina, Yifan He, and Ralph Grishman.
\newblock Personalized page rank for named entity disambiguation.
\newblock In \emph{NAACL}, 2015.

\bibitem[Peters et~al.(2018)Peters, Neumann, Iyyer, Gardner, Clark, Lee, and
  Zettlemoyer]{peters2018deep}
Matthew Peters, Mark Neumann, Mohit Iyyer, Matt Gardner, Christopher Clark,
  Kenton Lee, and Luke Zettlemoyer.
\newblock Deep contextualized word representations.
\newblock In \emph{NAACL}, 2018.

\bibitem[Radford et~al.(2018)Radford, Narasimhan, Salimans, and
  Sutskever]{radford2018improving}
Alec Radford, Karthik Narasimhan, Time Salimans, and Ilya Sutskever.
\newblock Improving language understanding with unsupervised learning.
\newblock Technical report, Technical report, OpenAI, 2018.

\bibitem[Radhakrishnan et~al.(2018)Radhakrishnan, Talukdar, and
  Varma]{Radhakrishnan2018-zj}
Priya Radhakrishnan, Partha Talukdar, and Vasudeva Varma.
\newblock {ELDEN}: Improved entity linking using densified knowledge graphs.
\newblock In \emph{NAACL}, 2018.

\bibitem[Raiman \& Raiman(2018)Raiman and Raiman]{Raiman2018-hm}
Jonathan Raiman and Olivier Raiman.
\newblock {DeepType}: Multilingual entity linking by neural type system
  evolution.
\newblock \emph{arXiv:1802.01021}, 2018.

\bibitem[Ratinov et~al.(2011)Ratinov, Roth, Downey, and
  Anderson]{ratinov2011local}
Lev Ratinov, Dan Roth, Doug Downey, and Mike Anderson.
\newblock Local and global algorithms for disambiguation to wikipedia.
\newblock In \emph{ACL}, 2011.

\bibitem[Riedel et~al.(2013)Riedel, Yao, McCallum, and
  Marlin]{riedel2013relation}
Sebastian Riedel, Limin Yao, Andrew McCallum, and Benjamin~M Marlin.
\newblock Relation extraction with matrix factorization and universal schemas.
\newblock In \emph{NAACL}, 2013.

\bibitem[Sil et~al.(2018)Sil, Kundu, Florian, and Hamza]{Sil2018-qo}
Avirup Sil, Gourab Kundu, Radu Florian, and Wael Hamza.
\newblock Neural {Cross-Lingual} entity linking.
\newblock In \emph{AAAI}, 2018.

\bibitem[Socher et~al.(2013)Socher, Chen, Manning, and Ng]{socher2013reasoning}
Richard Socher, Danqi Chen, Christopher~D Manning, and Andrew Ng.
\newblock Reasoning with neural tensor networks for knowledge base completion.
\newblock In \emph{NIPS}, 2013.

\bibitem[Sun et~al.(2015)Sun, Lin, Tang, Yang, Ji, and Wang]{sun2015modeling}
Yaming Sun, Lei Lin, Duyu Tang, Nan Yang, Zhenzhou Ji, and Xiaolong Wang.
\newblock Modeling mention, context and entity with neural networks for entity
  disambiguation.
\newblock In \emph{IJCAI}, 2015.

\bibitem[Toutanova et~al.(2015)Toutanova, Chen, Pantel, Poon, Choudhury, and
  Gamon]{toutanova2015representing}
Kristina Toutanova, Danqi Chen, Patrick Pantel, Hoifung Poon, Pallavi
  Choudhury, and Michael Gamon.
\newblock Representing text for joint embedding of text and knowledge bases.
\newblock In \emph{EMNLP}, 2015.

\bibitem[Toutanova et~al.(2016)Toutanova, Lin, Yih, Poon, and
  Quirk]{toutanova2016compositional}
Kristina Toutanova, Victoria Lin, Wen-tau Yih, Hoifung Poon, and Chris Quirk.
\newblock Compositional learning of embeddings for relation paths in knowledge
  base and text.
\newblock In \emph{ACL}, 2016.

\bibitem[Vaswani et~al.(2017)Vaswani, Shazeer, Parmar, Uszkoreit, Jones, Gomez,
  Kaiser, and Polosukhin]{vaswani2017attention}
Ashish Vaswani, Noam Shazeer, Niki Parmar, Jakob Uszkoreit, Llion Jones,
  Aidan~N Gomez, {\L}ukasz Kaiser, and Illia Polosukhin.
\newblock Attention is all you need.
\newblock In \emph{NIPS}, 2017.

\bibitem[Vilnis et~al.(2018)Vilnis, Li, Murty, and McCallum]{vilnis2018box}
Luke Vilnis, Xiang Li, Shikhar Murty, and Andrew McCallum.
\newblock Probabilistic embedding of knowledge graphs with box lattice
  measures.
\newblock In \emph{ACL}, 2018.

\bibitem[Wang et~al.(2018{\natexlab{a}})Wang, Cheng, Liu, and
  Liu]{wang2018additive}
Feng Wang, Jian Cheng, Weiyang Liu, and Haijun Liu.
\newblock Additive margin softmax for face verification.
\newblock \emph{IEEE Signal Processing Letters}, 25\penalty0 (7):\penalty0
  926--930, 2018{\natexlab{a}}.

\bibitem[Wang et~al.(2018{\natexlab{b}})Wang, Yan, and Wu]{wang2018multi}
Wei Wang, Ming Yan, and Chen Wu.
\newblock Multi-granularity hierarchical attention fusion networks for reading
  comprehension and question answering.
\newblock In \emph{ACL}, 2018{\natexlab{b}}.

\bibitem[Yaghoobzadeh \& Sch{\"u}tze(2015)Yaghoobzadeh and
  Sch{\"u}tze]{yaghoobzadeh2015corpus}
Yadollah Yaghoobzadeh and Hinrich Sch{\"u}tze.
\newblock Corpus-level fine-grained entity typing using contextual information.
\newblock In \emph{EMNLP}, 2015.

\bibitem[Yaghoobzadeh et~al.(2018)Yaghoobzadeh, Adel, and
  Sch{\"u}tze]{yaghoobzadeh2018corpus}
Yadollah Yaghoobzadeh, Heike Adel, and Hinrich Sch{\"u}tze.
\newblock Corpus-level fine-grained entity typing.
\newblock \emph{Journal of Artificial Intelligence Research}, 2018.

\bibitem[Yamada et~al.(2016)Yamada, Shindo, Takeda, and
  Takefuji]{yamada2016joint}
Ikuya Yamada, Hiroyuki Shindo, Hideaki Takeda, and Yoshiyasu Takefuji.
\newblock Joint learning of the embedding of words and entities for named
  entity disambiguation.
\newblock In \emph{CoNLL}, 2016.

\bibitem[Yamada et~al.(2017)Yamada, Shindo, Takeda, and
  Takefuji]{yamada-etal-2017-learning}
Ikuya Yamada, Hiroyuki Shindo, Hideaki Takeda, and Yoshiyasu Takefuji.
\newblock Learning distributed representations of texts and entities from
  knowledge base.
\newblock \emph{TACL}, 2017.

\bibitem[Yang et~al.(2014)Yang, Yih, He, Gao, and Deng]{yang2014embedding}
Bishan Yang, Wen-tau Yih, Xiaodong He, Jianfeng Gao, and Li~Deng.
\newblock Embedding entities and relations for learning and inference in
  knowledge bases.
\newblock In \emph{ICLR}, 2014.

\end{thebibliography}
\bibliographystyle{iclr2020_conference}

\end{document}


\maketitle

\appendix

\section{\ack Data}
\label{sec:data}

\subsection{Entity identifiers}

Unlike most entity-centric academic resources that use Freebase \citep{ling2012fine, riedel2013relation, yaghoobzadeh2015corpus}, we obtain our set of entities from Wikidata\footnote{https://www.wikidata.org} and Wikipedia.

One major limitation of Freebase is its reliance on fixed machine IDs (MIDs) for each entity. Because Freebase is now deprecated, and because MIDs changed with different versions of Freebase, it's difficult to obtain a unique identifier for a given entity. This means that large chunks of existing resources are no longer usable.
On the other hand, Wikidata is actively maintained, and so QIDs are the more stable choice going forward.

\subsection{Wikipedia dataset}


The data collection proceeds as follows:
\begin{enumerate}
    \item We take the 2018-10-22 dump of English Wikipedia.
    \item Obtain all sentences that have hyperlinks.
    \item For each sentence, replace all tokens in the hyperlink anchor text with a special \texttt{[MASK]} token, which anonymizes the entity in the context. 
    \item Tokenize the sentence using the wordpiece tokenizer of BERT. If the sentence is longer than 64 tokens, take the window of 64 tokens with the \texttt{[MASK]} token at the center.
    \item Label the ground truth entity of the sentence as the hyperlink target $e$.
\end{enumerate}

The final task is then to fill-in-the-blank, i.e. predict which entity goes in place of \texttt{[MASK]}.
Note in particular that we do not include the entity mention in our data, relying solely on the context instead; this differentiates our task from that of entity linking.

Overall, there are 5,165,638 entities and 112,352,037 (context, entity) pairs created by this process.
Table~\ref{tab:wikipedia_stats} shows statistics about number of examples per entity.

\section{Tasks}

\subsection{\ack training}

\begin{figure}[t]
    \centering
    \includegraphics[width=\columnwidth]{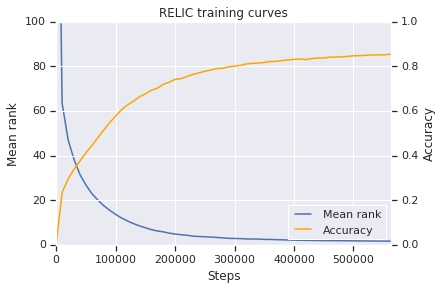}
    \caption{Mean rank and accuracy during \ack training.}
    \label{fig:training_curves}
\end{figure}

We initialize all non-BERT variables with a truncated normal distribution with standard deviation 0.02.
We train all parameters (including those of BERT) using the Adam optimizer \citep{kingma2014adam} with gradient norm clipped to 1.0. 
The learning rate follows a linear decay schedule with a maximum value of $5 \times 10^{-5}$ and a linear ramp-up of 10\% of the steps.

Figure~\ref{fig:training_curves} shows mean rank of the correct entity and accuracy during \ack training, both measured in the setting with batch negatives. We see that the model learns to fit the data reasonably well.

Table~\ref{tab:nearest_neighbors} shows examples of nearest neighbor entities. We see that similar entities are clustered as desired.

\begin{table*}[t]
    \centering
    \small
    \begin{tabular}{c|c|c|c}
    \toprule
    Valentina Tereshkova & rainbow & cancer & United States of America \\
    \midrule
Alexey Leonov & mirage & leukemia & Americans \\
Gherman Titov & snowflake & neoplasm & contiguous United States \\
Yuri Gagarin & mist & bone tumor & Eastern United States \\
Sergei Krikalev & halo & breast cancer & United Kingdom \\
Vladimir Komarov & sun dog & lymphoma & United States at the UCI Road World Championships \\
Pavel Popovich & glory & blood cancer & television in the United States \\
Scott Kelly & specular reflection & colorectal cancer & Federal Government of the United States \\
Sergei Korolev & Fata Morgana & lung cancer & Australian American \\
Andriyan Nikolayev & moonlight & carcinoma & Northwestern United States \\
Vladimir Dzhanibekov & darkness & skin cancer & North America \\    \bottomrule
    \end{tabular}
    \caption{Nearest neighbors of RELIC representations in cosine similarity.}
    \label{tab:nearest_neighbors}
\end{table*}

\subsection{Entity-level typing}

\paragraph{Data}
One mismatch between our data and those of FIGMENT and TypeNet is in the use of entity identifiers---external datasets are based on Freebase, which uses MIDs, while our data is based on Wikipedia and uses QIDs.
We resolve MIDs to QIDs using the correspondence in Wikidata when available. For FIGMENT, we get 93\% coverage, and for TypeNet we get 99\% coverage.

We assume type vocabulary $\mathcal{T}$, where $|\mathcal{T}| = 112$ for FIGMENT and $|\mathcal{T}| = 1077$ for TypeNet.

\paragraph{Model}
Our model is a simple feedforward network on top of our learned entity embedding.
Formally, given embedding $\mathbf{h}_e$ for entity $e$, we output a binary vector
\begin{equation}
    \mathbf{t}_e := \sigma(W_2 \cdot \operatorname{ReLU}(W_1 \frac{\mathbf{h}_e}{||\mathbf{h}_e||} + b_1) + b_2)
\end{equation}
where $W_1 \in \mathbb{R}^{d \times d_h}, b_1 \in \mathbb{R}^{d_h}, W_2 \in \mathbb{R}^{|\mathcal{T}| \times d_h}, b_2 \in \mathbb{R^{|\mathcal{T}|}}$ are learned parameters, and the output $\mathbf{t}_e$ are probabilities for each of the $|\mathcal{T}|$ types.
The parameters are learned by minimizing binary cross-entropy loss summed over all types.

Note that our embeddings are normalized to have norm 1 and are fixed during training.

\paragraph{Training}

For our model, we set $d_h = 500$ and apply dropout of 0.5 to the embedding and ReLU units.
We train with a batch size of 256 using Adam \citep{kingma2014adam} with a learning rate of $10^{-4}$ until convergence on the validation set.

The metrics used in FIGMENT are
\begin{enumerate}
    \item Precision@1: what percent of entities have their top-ranked type correctly labeled
    \item Accuracy: what percent of entities have all labels correct
    \item Micro F1: F1 aggregated over all binary decisions of entity and type
\end{enumerate}
as in \citet{yaghoobzadeh2015corpus}.

\subsection{TriviaQA}

\paragraph{Model} Our model for TriviaQA is exactly the same as that of \ack training, the only difference being that our inputs are questions instead of sentences with \texttt{[MASK]} token.
At inference time, we encode the sentence and retrieve the nearest neighbor entity out of the entire vocabulary of $\mathcal{E}$.

\paragraph{Training}
To convert TriviaQA into a classification problem, we first convert the provided text labels to our entity IDs.
We perform naive string matching on the text label to link it to a Wikipedia entity for training, skipping examples that we are unable to link.

During training, we tokenize a question $\mathbf{x}$ and feed it into our BERT context encoder $g(\mathbf{x})$. We adopt the same noise-contrastive loss with batch negatives as in \ack training.
On the entity side, we keep our \ack{} representations frozen, but fine-tune all other parameters.

We do two rounds of training, one with batch negatives and one with hard negatives.
First, starting from the \ack pretrained parameters, we fine-tune on the TriviaQA Wikipedia train set for 4000 steps with a batch size of 256 (about 20 epochs). We use the Adam optimizer \citep{kingma2014adam} with a learning rate of $5 \times 10^{-5}$ and the same learning rate schedule as in the \ack task.

To improve accuracy of classification, we also perform a second step of fine-tuning with \emph{hard negatives}, which we observed empirically to improve performance.
Specifically, we generate an approximate nearest neighbor index of our encodings, and during training, take the nearest encodings to $g(\mathbf{x})$ as negatives, and proceed as usual.
We use 128 negatives for all hard negative fine-tuning.

\bibliography{acl2019}
\bibliographystyle{acl_natbib}